\documentclass[lettersize,journal]{IEEEtran}
\usepackage{amsmath,amsfonts}
\usepackage{algorithmic}
\usepackage{algorithm}
\usepackage{array}
\usepackage[caption=false,font=normalsize,labelfont=sf,textfont=sf]{subfig}
\usepackage{textcomp}
\usepackage{stfloats}
\usepackage{url}
\usepackage{verbatim}
\usepackage{graphicx}
\usepackage{cite}
\usepackage{booktabs} 
\usepackage{tabularx}
\usepackage{booktabs}   
\usepackage{multirow}   
\usepackage{array}      
\usepackage{makecell}   
\usepackage{adjustbox}
\usepackage{amssymb}

\usepackage[table]{xcolor}
\definecolor{grey}{rgb}{0.5,0.5,0.5} 
\usepackage{pifont}
\usepackage{graphicx}
\usepackage{marvosym}

\begin{document}

\title{DailyArt: Discovering Articulation from Single Static Images via Latent Dynamics}


\author{
  \IEEEauthorblockN{
    Hang Zhang$^{1,2}$,
    Qijian Tian$^3$,
    Jingyu Gong$^1$,
    Daoguo Dong$^4$,
    Xuhong Wang$^2$,
    Yuan Xie$^1$,
    and Xin Tan$^{1,2,\textsuperscript{\Letter}}$
  }
  
  \IEEEauthorblockA{
    $^1$ East China Normal University 
  }
  \IEEEauthorblockA{
    $^2$ Shanghai AI Lab 
  }
  \IEEEauthorblockA{
    $^3$ Shanghai Jiao Tong University 
  }
  \IEEEauthorblockA{
    $^4$ Fudan University 
  }

  \thanks{\Letter~Xin Tan is the corresponding author.}
}

\markboth{Journal of \LaTeX\ Class Files,~Vol.~14, No.~8, August~2021}%
{Shell \MakeLowercase{\textit{et al.}}: A Sample Article Using IEEEtran.cls for IEEE Journals}


\maketitle

\begin{figure*}[t]
    \centering
    \includegraphics[width=\textwidth]{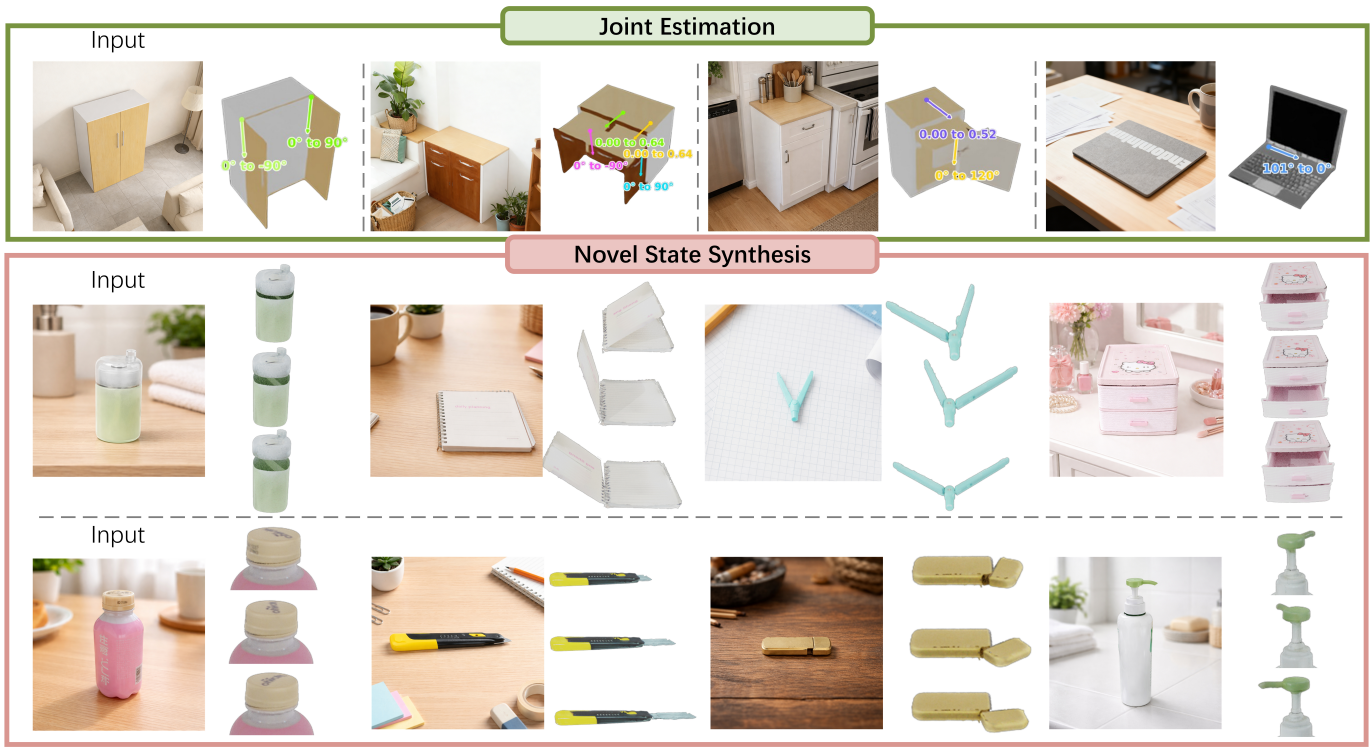}
    \caption{\textbf{Overview of DailyArt.}
    We propose a synthesis-mediated framework for articulated joint parameter estimation and controllable motion synthesis from a single static image. Given an input image, DailyArt first synthesizes a maximally articulated (opened) state to reveal hidden kinematic cues, which helps reduce 2D ambiguity. DailyArt (1) estimates joint parameters (type, axis, and motion range) from cross-state discrepancies, and (2) enables part-level articulated state synthesis.
    }
    \label{fig:teaser}
    \vspace{-6mm}
\end{figure*}
\begin{abstract} 
Articulated objects are essential for embodied AI and world models, yet inferring their kinematics from a single closed-state image remains challenging because crucial motion cues are often occluded. Existing methods either require multi-state observations or rely on explicit part priors, retrieval, or other auxiliary inputs that partially expose the structure to be inferred. In this work, we present DailyArt, which formulates articulated joint estimation from a single static image as a synthesis-mediated reasoning problem. Instead of directly regressing joints from a heavily occluded observation, DailyArt first synthesizes a maximally articulated opened state under the same camera view to expose articulation cues, and then estimates the full set of joint parameters from the discrepancy between the observed and synthesized states. Using a set-prediction formulation, DailyArt recovers all joints simultaneously without requiring object-specific templates, multi-view inputs, or explicit part annotations at test time. Taking estimated joints as conditions, the framework further supports part-level novel state synthesis as a downstream capability. Extensive experiments show that DailyArt achieves strong performance in articulated joint estimation and supports part-level novel state synthesis conditioned on joints. Project page is available at \url{https://rangooo123.github.io/DaliyArt.github.io/}.
\end{abstract}


\section{Introduction}
\label{sec:intro}


Articulated objects are not merely static props but interactive entities that are central to embodied AI and world models, where agents must perceive and manipulate their environments~\cite{black2024pi0,brohan2022rt,torne2024reconciling, tan2025towards,li2024flowbothd}. Humans can often infer how an object may be manipulated from a single glance, yet vision models struggle to recover the underlying kinematic structures (types, joint axes, and motion ranges), from a single closed-state view in which the relevant evidence is frequently occluded~\cite{duan2022survey,mandi2024real2code,li2024ag2manip}. This capability gap matters because actionable downstream applications require articulated assets with explicit joint parameters rather than just surface geometry alone~\cite{Mo2019partnetmobility,lian2025infinite,li2025triposg}.

Despite the growing interest in learned articulation inference, scaling articulated 3D assets remains challenging~\cite{mo2019partnet,xiang2025structured,li2024flowbothd}. Manual annotation provides accurate supervision but is labor-intensive and time-consuming~\cite{lai2025hunyuan3d,chen20253dtopia,tang2023dreamgaussian}. Learning-based pipelines reduce annotation cost, yet many of them rely on restrictive interfaces at test time~\cite{liu2025artgs,le2025articulateanything}. In particular, current methods typically assume access to part-level observations, using auxiliary inputs such as part masks, explicit part graphs, joint counts, or retrieval candidates from limited databases \cite{chen2024urdformer,le2025articulateanything,liu2024singapo}.

Existing approaches broadly fall into two paradigms, and both leave distinct limitations. One line of research relies on multi-state observations, extracting motion cues from image pairs or videos to cluster physical kinematics~\cite{liu2023paris,jiang2022ditto,yuan2025larm}. Although effective, this strategy shifts the burden to data collection, since an additional articulated state is rarely available at test time or in real-world situations~\cite{chen2024urdformer,mandi2024real2code}. Conversely, the other line of work stays in the single-state setting. Methods attempt to bypass this requirement by compensating with strong priors, such as retrieval, masks, or structural hints~\cite{gao2025partrm,le2025articulateanything,liu2024singapo,lu2025dreamart,li2025puppet,liu2025partfield}. However, this does not resolve the ambiguity of single-image articulation. Instead, it reduces the problem by injecting information that would otherwise need to be inferred. By introducing external specifications, these methods inadvertently pre-expose structural details that should be inferred~\cite{chen2024urdformer,mandi2024real2code}. As a result, these methods are often fragile when their assumptions do not hold at inference time, especially brittle under mismatched assumptions for novel objects or open-world diversity~\cite{chen2024urdformer,le2025articulateanything,liu2024singapo}.

%

\begin{figure}[t]
    \centering
    \includegraphics[width=\linewidth]{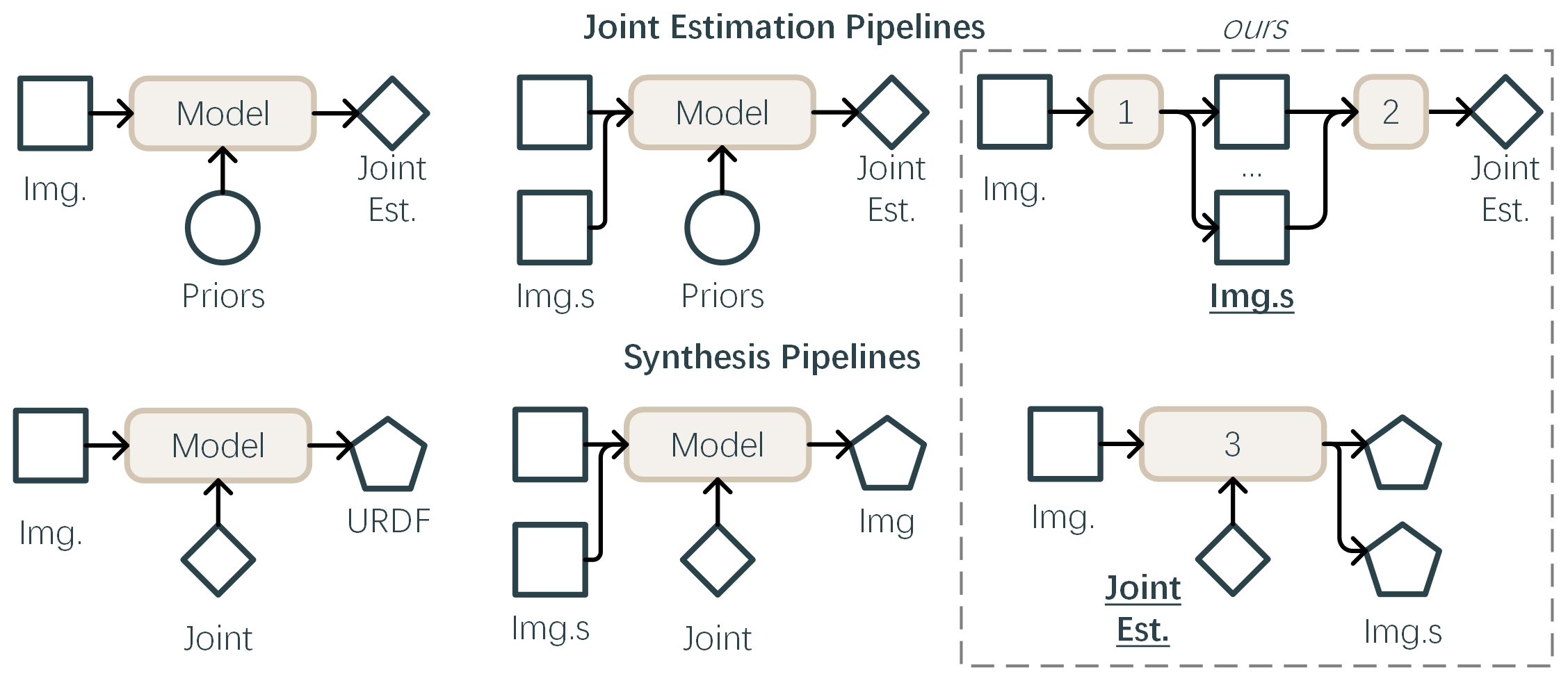}
    
    \vspace{-3mm}
    \caption{\textbf{Comparison of current pipelines (left \& mid) and our proposed DailyArt (right)}. Up: In joint estimation, existing pipelines leverage priors to guide the single image, multi-view or multi-state images. DailyArt generates novel opened-state images by encoding a state index into the image feature. The kinematic motion difference within dual-state images are directly compared and used to estimate the joints information. Down: For showing novel states of an object, some methods import URDF files into simulators. DailyArt uses both the input image and estimated joints to synthesis multiple states of every kinematic joint.}
    \label{fig:intro_comparison}
    \vspace{-3mm}
\end{figure}

We identify the core challenge in estimating joints from a static closed-state image. When kinematic cues are occluded beneath the surface, one observation may support several plausible joint interpretations. Existing methods typically narrow this space with explicit annotations or structural priors. Yet such priors are often unavailable for novel objects, and even segmentation-based cues can fail movable parts and the static body share nearly indistinguishable appearance in the closed state. We therefore replace these auxiliary priors with autonomous articulated state synthesis. The intuition is simple. Akin to human reason about joints via first imaging how parts might move eventually, we argue that synthesized dual-state evidence offers a promising play for joint estimation. 
This perspective also reveals a prior-dependency paradox in current pipelines. As shown in Fig.~\ref{fig:intro_comparison}, existing generative models~\cite{gao2025partrm,li2025puppet,lu2025dreamart} usually require interactive guidance to indicate which part should move. Kinematic predictors, in turn, often require the number of parts or the topology to be specified in advance. This forms a circular dependency: synthesis is needed to expose motion evidence for joint estimation, but existing synthesis pipelines depends on the very part-level information that joint estimation is supposed to discover. To break this loop, we propose synthesizing a maximally articulated state without part-level guidance, which also benefits the later estimation by contributing all potentially movable parts. This design calls for a unified redesigned pipeline that does not depend on topology assumptions during synthesis and does not require part annotations during inference.


Motivated by this insight, rather than presenting a generic articulated object generation framework, we focus on articulated joint estimation from a single static image and formulate it as a \textit{synthesis-mediated reasoning} problem. We introduce \textbf{DailyArt}, a framework that separates target-state synthesis from downstream joint estimation. Instead of predicting joints directly from a heavily occluded closed-state observation, DailyArt first synthesizes a physically plausible opened state, and then estimates kinematics from the discrepancy between the observed and synthesized states.


DailyArt follows a three-stage pipeline centered on articulated joint estimation. In Stage I, we train a state synthesis model that maps a single closed-state image ($t=0$) to a maximally articulated state ($t=1$). This stage is designed to expose articulation cues rather than to provide part-level control. In Stage II, we lift the synthesized image pair $(\mathbf{I}_{0}, \mathbf{\hat{I}}_{1})$ into dense, confidence-aware 3D point maps to reduce image-space ambiguity. A set-prediction formulation then recovers all joint parameters, including joint types, pivot origins, axis directions, and motion limits in object-centered world coordinates, within a single forward pass. In Stage III, we feed the estimated joints back into the synthesis backbone as explicit conditions, enabling part-level articulation synthesis. In this sense, the final stage is a downstream capability built on top of the joint reasoning pipeline, rather than the primary target of the method.



In summary, DailyArt formulates articulated joint estimation from a single static image as a synthesis-mediated reasoning problem. Our core contributions are:
\begin{itemize}
\item \textbf{A synthesis-mediated formulation for articulated joint estimation.} We formulate full articulated joint estimation from a single static image as a synthesis-mediated reasoning problem, without requiring priors such as CAD models, multi-view inputs, or explicit part annotations.
\item \textbf{Joint-conditioned novel state synthesis.} We further show that the estimated joints can be fed back into the synthesis backbone to enable novel articulation state synthesis for individual movable parts. This makes the recovered kinematic parameters directly usable for controllable image-space articulation beyond joint estimation.
\end{itemize}

\section{Related Work}
\label{sec:related}

\subsection{Multi-State Reconstruction Methods}
An early standard way to make articulation estimation well-posed is to observe motion across states~\cite{wei2022self,mu2021sdf,jain2022distributional,patil2023rosi,song2024reacto,deng2024articulate}. Such as PARIS~\cite{liu2023paris} and ArticulateGS~\cite{guo2025articulatedgs} aligning reconstructions across articulation states, many pipelines leverage multi-state observations (image pairs, videos, or induced interactions) to expose moving parts and recover kinematics with explicit cross-state evidence \cite{liu2023paris,weng2024neural,jin2025lvsm}. Multi-view capture further strengthens geometric constraints and enables more accurate joint localization and axis estimation \cite{park2021nerfies,park2021hypernerf}. Recent feed-forward models scale this principle by taking sparse views from two distinct articulation state pairs (e.g., rest and limit) as inference inputs to regress deformation and joint parameters in a single pass~\cite{yuan2025larm,kawana2023detection,che2024op}. Related works in robotics and interaction learning similarly rely on generative~\cite{mandi2024real2code,cao2025physx} or language priors~\cite{chen2025freeart3d} to reveal articulation cues and learn kinematic structure. While SINGAPO~\cite{liu2024singapo} and MeshArt~\cite{gao2025meshart} predict graphic trees and retrieve articulated parts, Articulate-Anything~\cite{le2025articulateanything} reforms the prior requirements into LLM reasoning on object videos and PhysX-Anything~\cite{cao2025physx} scales the physical structure process into simulation engines using VLMs. DailyArt targets a different input interface: a single closed-state RGB image at test time. Instead of requiring an additional state or interaction, we synthesize a plausible target articulated state to construct cross-state evidence under the same camera viewpoints, and then infer kinematics from the induced discrepancy.

\subsection{Single-Image Methods with Priors}
When only a single image is available, articulation inference is typically regularized by priors. One line predicts articulated representations (e.g., URDF-like parameters) directly from images by learning category-level structural assumptions~\cite{chen2024single,chen2024urdformer,fan2017point,cao2025physx}. Another line introduces external semantic specifications via retrieval or tool-use pipelines. Foundation models, like Articulate-Anything~\cite{le2025articulateanything}, propose graphic part structure and joint hypotheses, which are then matched to databases or procedural templates~\cite{le2025articulateanything,liu2024singapo}. In a similar spirit, single-image controllable generation methods~\cite{poole2022dreamfusion,zhang2023controlnet,jiang2022opd} synthesize articulated parts under additional constraints such as part masks, motion prompts, or category-level structure priors~\cite{lu2025dreamart,liu2024singapo,pan2023draggan,li2024dragapart,tu2025dreamo} or with pseudo multi-view constraints~\cite{liu2024one,gao2025partrm,li2026monoart}. These approaches~\cite{pan2023draggan,chen2024urdformer} demonstrate the value of priors in reducing ambiguity, but they also expand the test-time input contract (masks/ graphs/ prompts/ part counts) and can be brittle when priors are incomplete or mismatched across open-world objects without human adjustments \cite{long2024wonder3d,liu2024singapo,chen2024urdformer}. In contrast, DailyArt keeps inference image-only (no masks, graphs, prompts, or manual declarations of part counts/joint types). We instead construct motion evidence through synthesis-first reasoning, converting under constrained single image regression into cross-state estimation.

\begin{figure}[t]
    \centering
    \includegraphics[width=0.95\linewidth]{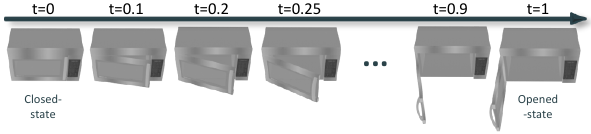}
    \vspace{-3mm}
    \caption{\textbf{Kinematic states of an articulated object}. We define $t=0$ as the closed state where the part is closed or remains inactivated, and define $t=1$ as the opened state where the part reaches the maximum articulated limit. And the motion index $t=t^{\prime}$ is a condition to describe novel states of parts somewhere within the motion range.}
    \label{fig:dataflow}
    \vspace{-3mm}
\end{figure}

\subsection{Generative Methods with Kinematic Clues}
Generative models provide an alternative source of motion cues when observations are limited. Recent work synthesizes articulated motion or state change from single images or interactive controls, ranging from part-level controllable generation to motion prior learning from large-scale video data \cite{yin2023dragnuwa,pan2023draggan,lu2025dreamart,li2025puppet,gao2025partrm}. In parallel, articulated 3D generation explores structured representations that disentangle geometry and articulation to improve realism and controllability~\cite{mo2019partnet,chen2025artilatent,chen2025freeart3d}. More broadly, progress in 3D generative priors and supervision resources underpins these directions, including score-distillation-based 3D synthesis and diffusion backbones~\cite{sun2021loftr,poole2022dreamfusion,lin2023magic3d,liu2023zero,long2024wonder3d}, as well as large 3D asset corpora and strong pre-trained visual encoders \cite{hessel2021clipscore,deitke2023objaverse,deitke2023objaversexl,oquab2023dinov2,kirillov2023sam}. 


\begin{figure*}[t]
    \centering
    \includegraphics[width=\textwidth]{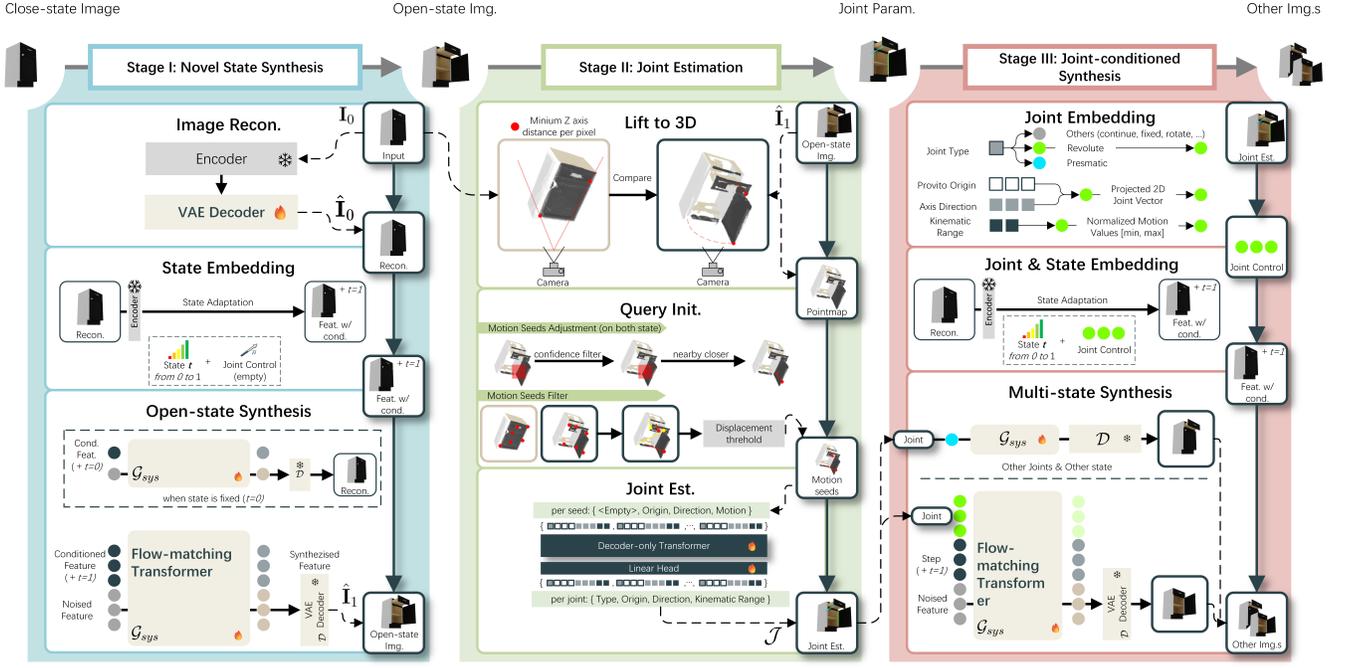}
    \caption{{\bf Overview of the DailyArt Framework.} Given a single closed-state image $\mathbf{I}_0$, DailyArt adopts a three-stage paradigm to estimate joints and synthesis images. The input is processed by a prior-free novel state synthesis (Stage I) to an opened-state $\hat{\mathbf{I}}_{1}$, revealing occluded motion evidence. For joint estimation (Stage II), the input and synthesized states are lifted into 3D point-maps as $(\mathcal{P}_0, \mathcal{P}_{1})$ to estimate a set of joint parameters $\hat{\mathcal{J}}$. Having estimated joints, DailyArt could extend the novel state synthesis as joint-conditioned synthesis (Stage III) from the input $\mathbf{I}_0$ to novel states related to different joints $\hat{\mathbf{I}}_{t=t^{\prime}, \mathcal{J}}$. DailyArt performs outputs without part annotations or priors.}
    \label{fig:mainpipeline}
    \vspace{-3mm}
\end{figure*}

\section{Method}
\label{sec:method}
\subsection{Overview}
Given a single closed-state image $\mathbf{I}_{0} \in \mathbb{R}^{C \times H \times W}$, DailyArt is expected to estimate a set of $N$ articulated joints $\mathcal{J}= \{\mathbf{J}_n\}_{n=1}^N$ ($N$ is the ground-truth joint count for one object, later $K$ as estimated number). Each articulated joint $\mathbf{J}_n$ includes type value ${\tau}_n \in \{0,1,2,\dots,6\}$ (fixed, rotate, revolute, continuous, prismatic, etc.) to fit the annotations in baseline URDF files~\cite{chen2024urdformer,liu2024singapo,le2025articulateanything,cao2025physx}, where we annotate origin position vector $\mathbf{o}_n$ $\in\mathbb{R}^{3}$, axis direction vector $\mathbf{a}_n$ $\in\mathbb{R}^{3}$ and motion vector $\boldsymbol{\vartheta}_n$$=(m_n^{\min}, m_n^{\max})$ $\in\mathbb{R}^{2}$. 
Based on the estimated joints, DailyArt can further synthesize joint-conditioned articulation sequences for individual movable parts as $\{\mathbf{\hat{I}}\}_{t\in T, n \in N}$ $\in$ $\mathbb{R}^{T \times N \times C \times H \times W}$, where $T$ is the state sequence length. As shown in Fig.~\ref{fig:dataflow}, a state sequence depicts motion at a single articulated joint. The corresponding kinematic part is gradually opened from the closed-state (annotated as $t=0$) until the kinematic part reaches the motion limit at the opened-state ($t=1$). DailyArt reformulates single-image articulation estimation and novel-state synthesis from a single image into multiple progressive stages. As illustrated in Fig.~\ref{fig:mainpipeline}, the three-stage framework is built on novel state synthesis (\ref{sec:method_stage1}), joint estimation (\ref{sec:method_stage2}), and kinematic controlled synthesis (\ref{sec:method_stage3}).

\subsection{Stage I: Novel State Synthesis}
\label{sec:method_stage1}
Stage~I synthesizes novel articulated state transitions conditioned on a scalar index $t$, representing the extent of kinematic motion, and an empty condition $\varnothing$ that serves as a placeholder for the explicit joint articulation introduced later in Stage~III. The synthesis process is expected to produce state transitions that strictly preserve the identity and geometry of the original object, because high visual consistency in 3D is necessary for the dense dual-state motion comparison in Stage~II. Although articulated motions resemble short video sequences, adopting a standard video diffusion model does not align with our constraints. Diffusion models typically require precise structural priors or external guidance to maintain temporal consistency. Our configuration intentionally restricts access to these priors and requires the model to synthesize state transitions purely from the latent semantics of the single input image.

\paragraph{Synthesis Backbone}
We build the synthesis backbone from a frozen image encoder $\mathcal{E}$ and a learnable decoder $\mathcal{D}$ to achieve a reconstruction performance first. The encoder $\mathcal{E}$ adopts DINOv2 to extract semantically registered image features from the input $\mathbf{I}_0$. Given the patchified token sequence $\text{tok}=\mathcal{E}(\mathbf{I}_0)$, we construct a VAE-based decoder that maps the semantic latent space back to pixels. Formally, the reconstruction branch outputs $\hat{\mathbf{I}}_{0} = \mathcal{D}(\text{tok})$.

\paragraph{State-conditioned Synthesis}
On top of the synthesis backbone, Stage~I synthesizes kinematic states conditioned on a scalar kinematic index $t$. We encode $t$ with a sinusoidal embedding and map it to the same latent dimension $\mathcal{T}$ as the image tokens. The encoded image tokens and state embedding are fused through Adaptive Layer Normalization (AdaLN),
\begin{equation}
\text{AdaLN}(\text{tok}, \mathcal{T}) = \gamma(\mathcal{T})\cdot \text{LayerNorm}(\text{tok}) + \beta(\mathcal{T}),
\end{equation}
where $\gamma(\mathcal{T})$ and $\beta(\mathcal{T})$ are scale and shift parameters regressed by an MLP from $\mathcal{T}$. Importantly, Stage~I does not yet specify which joint should move. Instead, it learns to generate a maximally articulated state that exposes as much articulation evidence as possible, while the joint condition is kept empty as a placeholder. This choice is deliberate: Stage~I is designed for articulation cue discovery rather than precise component-level control. Stage I novel state synthesis branch towards the opened-state is therefore written as
\begin{equation}
\hat{\mathbf{I}}_{t=1} = f_{\text{Stage I}}((\mathbf{I}_0), t=1) = \mathcal{D}\big(\mathcal{S}(\mathcal{E}(\mathbf{I}_0), t=1, \varnothing)\big),
\end{equation}
where $\mathcal{S}$ denotes the state adaptation module and $\varnothing$ denotes the empty joint condition. $t=1$ represents the model produces the maximally articulated state $\hat{\mathbf{I}}_{1}$ used by Stage~II for joint estimation. Additionally, intermediate values of $t$ $\in (0,1)$ (not included in Stage~I) correspond to partially articulated states and are treated as a natural extension of the same synthesis mechanism later in Stage~III.

\subsection{Stage II: 3D-Aware Joint Estimation}
\label{sec:method_stage2}
To mitigate the ambiguity of 2D observations, particularly when they are weakly visible in the input view, Stage~II leverages the cross-state discrepancy between the closed-state $\mathbf{I}_0$ (input) and the synthesized opened-state $\hat{\mathbf{I}}_{1}$ (maximally articulated). Rather than estimating joints directly from 2D appearance, we first lift the image pair $(\mathbf{I}_0, \hat{\mathbf{I}}_{1})$ into dense 3D point-maps $\mathcal{P}_0, \mathcal{P}_{1} \in \mathbb{R}^{H \times W \times 3}$ using a pre-trained Vision Geometry Transformer (VGGT) \cite{wang2025vggt}. This allows us to reason joint axes in world coordinates without camera extrinsics:
\begin{equation}
(\mathcal{P}_0,\mathcal{C}_0) = \Phi(\mathbf{I}_0), \qquad
(\mathcal{P}_1,\mathcal{C}_1) = \Phi(\hat{\mathbf{I}}_1),
\end{equation}
where $\mathcal{C}_0,\mathcal{C}_1 \in \mathbb{R}^{H \times W}$ are the corresponding confidence maps. The direct comparison between point clouds allows Stage~II to further reason about articulation in world coordinates without committing early to explicit part decomposition.

\paragraph{Motion Seed Extraction \& Filtering} We compare and compute the per-point 3D displacement $\Delta \mathcal{P}$. As shown in Fig.~\ref{fig:mainpipeline}, a motion seed is retained per pixel at the image coordinate $u$ by the paired 3D positions $[\mathcal{P}_0(u), \mathcal{P}_{1}(u)]$ where the same coordinate is observed among dual states (close and open). These points are initalized based on the minimum distance from the negative Z axis to the camera centre. Next, to handle the errors in the initalized seed coordinates, we retain motion seeds whose displacement magnitude falls within two steps. (1) 3D adjustment: We adjust the motion seeds that are spatially inconsistent with observable articulation. To remove points with low geometric confidence (often manifesting as 'white ribbon' artifacts or background noise in VGGT outputs), we set the confidence threshold to 0.85 as $\min$ $(\mathcal{C}_0$ $(u)$,$\mathcal{C}_1(u))$$>\text{conf}=0.85$. For both states, every seed is checked again as the closed point; otherwise, the seed is adjusted towards the closer (to the camera centre) point nearby. (2) Displacement filtering: Let $d(u)=\|\Delta \mathcal{P}(u)\|$ denote the displacement magnitude of each candidate motion seed. We rank all candidate seeds by $d(u)$ and discard both extremes: the shortest $15\%$ of seeds, which are often dominated by minor geometric noise, and the longest $20\%$ of seeds, which tend to correspond to unstable or overly large diagonal motions. We retain only the middle range of seeds. These percentile thresholds are determined empirically from the displacement statistics of the training set. This filtering process is intentionally non-learned.

\paragraph{Multiple Joint Estimation}
The filtered motion seeds initialize a set of joint queries $\mathcal{Q}$, each embedded with its 3D position. $\mathcal{Q}$ is concatenated with image-pair features after $\mathcal{E}$ and processed by a transformer-based estimator. Stage~II estimates joints $\hat{\mathcal{J}}$ $\in \mathbb{R}^{K \times 9}$, where $K$ is a pre-set upper bound on the number of articulated joints (set as 16, larger than the maximum number of objects' joints in the dataset). Thus, the Stage II process is written as
\begin{equation}
\hat{\mathcal{J}} = f_{\text{Stage II}}((\mathbf{I}_0), \hat{\mathbf{I}}_{1}).
\end{equation}

During training, we use Hungarian matching~\cite{kuhn1955hungarian} to assign each ground-truth joint to at most one predicted hypothesis, sorted on the predicted type. The matched hypotheses are supervised as articulated joints, while the unmatched hypotheses are optimized toward the fixed part and treated as unused slots. During inference, we retain only predictions whose confidence exceeds a threshold and whose predicted type is not fixed ('fixed' indicates the static base of the object or handles).

\subsection{Stage III: Joint-conditioned State Synthesis}
\label{sec:method_stage3}
Stage~III extends the synthesis model in Stage~I by introducing an explicit joint condition, describing one part-level articulation. This design turns the prior-free state synthesis branch into a controllable rendering module that can visualize the estimated articulation on the input object. The key distinction from Stage I is that the model synthesizes the fully opened image by using only the scalar state index $t$, whereas Stage~III specifies which articulated joint plays as a condition.

Given a selected predicted joint $\hat{\mathcal{J}}_k$ ((where the kinematic type $\hat{\tau}_k \ge 1$)) from the estimated set $\hat{\mathcal{J}}$ and a target articulation state $t' \in [0,1]$, Stage~III synthesizes the corresponding component-level articulated image as
\begin{equation}
\hat{\mathbf{I}}_{t=t^\prime, \hat{\mathcal{J}}} = = f_{\text{Stage II}}(\mathbf{I}_0), t=t^\prime, \hat{\mathcal{J}})\big) = \mathcal{D}\big(\mathcal{S}(\mathcal{E}(\mathbf{I}_0), t=t^\prime, \hat{J})\big),
\end{equation}
where the articulation state $t^\prime$ is not a specific physical degree or distance, yet a value within the closed-state $t=0$ and opened-state $t=1$. The synthesis module then generates image $\tilde{\mathbf{I}}_t$ additionally constrained by the estimated $\hat{J}$. As a result, the generated articulation becomes visually consistent with the recovered joint type, axis direction, pivot location, and motion range.

\subsection{Training Schedule and Loss Objectives}
\label{sec:method_training}
DailyArt is trained progressively. We first warm up the reconstruction-aligned backbone in Stage~I, then optimize state-conditioned synthesis on the same backbone, next train the joint estimator in Stage~II from the input and synthesized opened-state, and finally specialize the synthesis backbone in Stage~III with explicit joint conditioning.

\paragraph{Stage~I Pixel-level Loss} To reconstruct per image $\hat{\mathbf{I}}$ from the input $\mathbf{I}$, we train the decoder $\mathcal{D}$ to align with the frozen encoder $\mathcal{E}$ with a combination of L1 loss $\lambda_{\text{L1}}=0.9$ and the perceptual loss (LPIPS) $\lambda_{\text{LPIPS}}=0.1$:
\begin{equation}
    \mathcal{L}_{\text{rec}} = \lambda_{\text{L1}} \| \hat{\mathbf{I}} - \mathbf{I} \|_1 + \lambda_{\text{LPIPS}} \mathcal{L}_{\text{LPIPS}}(\hat{\mathbf{I}}, \mathbf{I}).
\end{equation}

With $\mathcal{E}$ and $\mathcal{D}$ frozen after pre-trained loss $\mathcal{L}_{\text{rec}} \leq 1e-6$ , we optimize $\mathcal{S}$ for state-conditioned synthesis. Given a target state index $t^\prime=1$, the synthesized image is supervised in image space, for one input, we have
\begin{equation}
\mathcal{L}_{\text{I}} =
\left\|
\mathcal{D}\big(\mathcal{S}(\mathcal{E}(\mathbf{I}_0), t=1, \varnothing)\big)
-
\mathbf{I}_{t=1}
\right\|_2^2.
\end{equation}

\paragraph{Stage~II Joint Estimation Loss}
For joint estimation, Stage~II takes the image pair $(\mathbf{I}_0,\hat{\mathbf{I}}_1)$ from Stage~I and predicts a set of $K$ joint hypotheses
$\hat{\mathcal{J}}=\{\hat{\mathbf{J}}_k\}_{k=1}^{K}$,
where each hypothesis is parameterized as
$\hat{\mathbf{J}}_k=(\hat{\tau}_k,\hat{\mathbf{o}}_k,\hat{\mathbf{a}}_k,\hat{\boldsymbol{\vartheta}}_k)$,
including the predicted joint type, pivot origin, axis direction, and motion range.
The ground-truth joint set is denoted as
$\mathcal{J}=\{\mathbf{J}_n\}_{n=1}^{N}$, with $N \leq K$. During training, we use Hungarian matching~\cite{kuhn1955hungarian} to obtain an injective assignment $\sigma(n)$ from each ground-truth joint $\mathbf{J}_n$ to one predicted hypothesis $\hat{\mathbf{J}}_{\sigma(n)}$. Matched predictions are supervised as articulated joints, while unmatched predictions are assigned to the \textit{fixed} class.

We first define a slot-wise classification target $\tau_k$ for each predicted hypothesis, where $\tau_k$ is the ground-truth joint type if $k=\sigma(n)$ for some $n$, and \textit{fixed} otherwise. The classification loss is defined to measure the joint types as
\begin{equation}
\mathcal{L}_{\text{cls}}=
\frac{1}{K}\sum_{k=1}^{K}\mathcal{L}_{\text{CE}}(\tau_k,\hat{\tau}_k).
\end{equation}

For each matched pair $(\mathbf{J}_n,\hat{\mathbf{J}}_{\sigma(n)})$, we optimize the joint pivot, axis direction, and motion range by
\begin{equation}
\begin{aligned}
\mathcal{L}_{\text{joint}}(\mathbf{J}_n,\hat{\mathbf{J}}_{\sigma(n)})
&=
\|\mathbf{o}_n-\hat{\mathbf{o}}_{\sigma(n)}\|_2^2
+
1-\cos(\mathbf{a}_n,\hat{\mathbf{a}}_{\sigma(n)}) \\
&+
\|\boldsymbol{\vartheta}_n-\hat{\boldsymbol{\vartheta}}_{\sigma(n)}\|_2^2.
\end{aligned}
\end{equation}

The overall Stage~II objective is
\begin{equation}
\mathcal{L}_{\text{II}}
=
\mathcal{L}_{\text{cls}}
+
\lambda_{\text{reg}}
\frac{1}{N}\sum_{n=1}^{N}
\mathcal{L}_{\text{joint}}(\mathbf{J}_n,\hat{\mathbf{J}}_{\sigma(n)}),
\end{equation} 
where $\boldsymbol{\vartheta}_n$ denotes the motion range parameters. Since revolute and prismatic joints are measured in different physical units, we normalize motion ranges to $[0,2]$ for more balanced regression (mapping the $[-360^{\circ},360^{\circ}]$). We map related values back for evaluations.

\paragraph{Stage III Joint-conditioned Synthesis Loss}
Given an articulation state $t=t'$ and the ground-truth joint condition $\mathcal{J}$ (training-only), the Stage~III output is supervised against the corresponding target image:
\begin{equation}
\mathcal{L}_{\text{III}} =
\left\|
\mathcal{D}\big(\mathcal{S}(\mathcal{E}(\mathbf{I}_0), t=t', \mathcal{J})\big)
-
\mathbf{I}_{t'}
\right\|_2^2.
\end{equation}

\paragraph{Inference Pipeline.}
At test time, the pipeline operates in a feed-forward manner progressively. Given a single image $\mathbf{I}_0$, Stage~I first synthesizes the maximally articulated state $\hat{\mathbf{I}}_{1}$. Stage~II then lifts the paired results $(\mathbf{I}_0,\hat{\mathbf{I}}_{1})$ from Stage~I into 3D and predicts the joint set $\hat{\mathcal{J}}$. Stage~III reuses the same synthesis backbone with the estimated results$\hat{\mathcal{J}}$ from Stage~II as an explicit condition to generate the target articulated image $\tilde{\mathbf{I}}_t$ at any desired state $t$.


\section{Experiments}
\label{sec:exp}

\subsection{Experimental Setup}
\label{sec:setup}
\paragraph{Baselines} Since DailyArt takes a single image as input and synthesizes novel state images and estimates joints, we compare two groups of evaluations (see Table~\ref{tab:exp_baselineinputs}) and provide extra information on baselines if required (i.e. priors or part masks). (1) Novel State Synthesis (Image Output): In this task, we evaluate DailyArt in novel state synthesis compared with recent state-of-the-art approaches: DragAPart~\cite{li2024dragapart}, PartRM~\cite{gao2025partrm}, Puppet -Master~\cite{li2025puppet} and LARM~\cite{yuan2025larm}. (2) Articulated Joint Estimation: DailyArt estimates joint parameters $\mathcal{J}$, compared with methods output URDF or json files with clear joint annotations: URDFormer~\cite{chen2024urdformer}, Singapo~\cite{liu2024singapo}, ArticulateAnything~\cite{le2025articulateanything} and PhysX-Anything~\cite{cao2025physx}. 

\begin{table}[t]
\centering
\caption{\textbf{Task Level Comparisons.} We disclose the input modalities and extra requirements for each baseline. \textbf{DailyArt} is the only method that enables both high-fidelity synthesis and precise kinematic estimation from a \textbf{single static image} without requiring interaction, retrieval, or language prompts.}
\label{tab:exp_baselineinputs}
\vspace{-3mm}
\setlength{\tabcolsep}{3pt}
\resizebox{0.95\linewidth}{!}{
\begin{tabular}{l|c|ccc}
\toprule
\textbf{Method} & \textbf{Single Image} & \textbf{Extra Priors} & \textbf{Interaction} & \textbf{Multi-State} \\
\midrule
\multicolumn{5}{c}{\cellcolor{gray!5}\textbf{\textit{Novel State Synthesis Baselines}}} \\
DragAPart & \ding{51} & -- & Drag Points & -- \\
PartRM & -- & Drag from Multi-state Masks & -- & Zero123+ \\
Puppet-Master & \ding{51} & -- & Drag Points & -- \\
LARM & -- & Multi-views & Camera Position & As Inputs \\
\midrule
\multicolumn{5}{c}{\cellcolor{gray!5}\textbf{\textit{Joint Estimation Baselines}}} \\
URDFormer & \ding{51} & Part Annotations & Human Adjustment & -- \\
Singapo & \ding{51} & GPT-4o & Data Retrial & -- \\
Articulate-Anything & - & LLM Prior & Data Retrial & Dense Video \\
PhysX-Anything & \ding{51} & QWen & Engine-based & -- \\
\midrule
\rowcolor{grey!10} \textbf{DailyArt (Ours)} & \ding{51} & -- & -- & -- \\
\bottomrule
\end{tabular}
}
\vspace{-3mm}
\end{table}

\paragraph{Dataset} We evaluate DailyArt and the baselines on PartNet Mobility~\cite{xiang2020sapien}, which serves as a benchmark for fine-grained articulated objects. Following~\cite{gao2025partrm,li2025puppet,liu2024singapo,le2025articulateanything}, we render 2.7k training samples from categories including Dishwasher, Folding Chair, Glasses, Laptop, Microwave, Oven, Printer, Refrigerator, Storage Furniture, Table, Suitcase, and Trashcan, and use another 347 objects for testing under the same train-test split in Blender. To expose the model to a broader range of 3D objects, we pre-train the decoder $\mathcal{D}$ on images from Objaverse-XL~\cite{deitke2023objaversexl}, excluding articulated objects. We further evaluate zero-shot performance on novel state synthesis and joint estimation by using real-world objects in the AKB-48 dataset~\cite{liu2022akb}, without any training.

\begin{figure*}[t]
    \centering
    \includegraphics[width=0.95\linewidth]{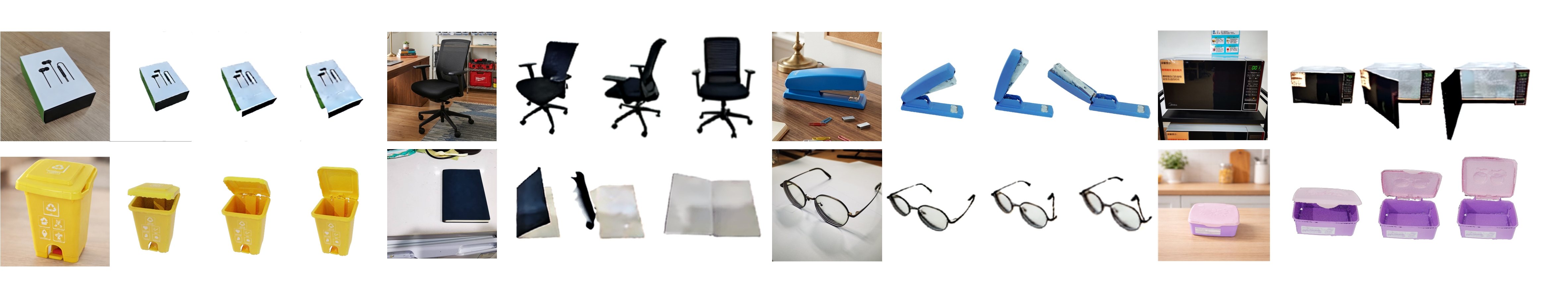} 
    \caption{{\bf Unseen object test results.} We test DailyArt performance on unseen objects. The images are segmented with a transparent background as inputs. The results demonstrate that DailyArt can handle such inputs and synthesise novel states.}
    \label{fig:conclu_failure}
    \vspace{-3mm}
\end{figure*}

\paragraph{Metrics} For \textbf{Novel State Synthesis}, we report PSNR, SSIM~\cite{wang2004image}, and LPIPS~\cite{zhang2018unreasonable} to evaluate synthesis images with ground truth, and CLIP-T (CLIP Score)~\cite{hessel2021clipscore} and FVD (Fréchet Video Distance)~\cite{heusel2017gans} to verify novel-state semantic alignments. For \textbf{Joint Estimation}, we adopt defined metrics from Articulate-Anything~\cite{le2025articulateanything} on axis angle error, origin point distance, motion range and axis direction. To assess the overall reliability of the system, we report a composite Success Rate, the success rate reported as metrics under 0.25 radians, 0.15, 0.3, 0.3 for axis angle error, axis origin error, motion error and direction error, respectively.

\paragraph{Implementation Details} We employ DINOv2 (ViT-L/14)~\cite{oquab2023dinov2} as our primary visual encoder. All modules are implemented in PyTorch and optimized using AdamW ($\beta_1=0.9, \beta_2=0.95$, weight decay $0.05$). We adopt a decoupled training schedule: Stage I is (Representation Alignment) trained for 20k epochs using AdamW with a batch size of 32 and an initial learning rate (LR) of $2 \times 10^{-5}$. The alignment is supervised by a combination of $L_1$ reconstruction loss ($\lambda_{L1}=0.9$) and a VGG-based perceptual loss ($\lambda_{perc}=0.1$). Training stops once $L_1$ loss drops under $1e-6$. Then, Stage II (Joint Estimation) is trained on paired $(\mathbf{I}_0, \hat{\mathbf{I}}_1)$ samples for 500 epochs with an initial LR of $2 \times 10^{-5}$ while keeping the DINOv2 backbone frozen. The best Stage II checkpoint is selected based on validation Overall SR. Stage III is initialized from the Stage I backbone and trained for 1k epochs with grounth truth joints as conditioning signals (estimated joints from Stage II at inference time). Unless otherwise specified, all reported results use the best validation checkpoint of each stage. Overall, the training is conducted on a cluster of 8 NVIDIA H200 GPUs (140GB) with a batch size of 128. Images are resized to $224 \times 224$. At test time, a single forward pass through our pipeline takes 0.45s on a single GPU (280ms for Stage 1 synthesis and 170ms for Stage 2 estimation), making it suitable for interactive applications.

\subsection{Main Results}
\label{sec:main_results}
We follow the original protocols of all baselines when preparing their required priors. For methods that assume additional structural inputs, we provide those priors accordingly, including ground-truth priors when required by the original setting. For joint estimation, we evaluate each predicted attribute under the URDF file parameterization. All quantitative results are averaged over 5 runs with random seeds 42, 43, 2024, 20525, and 2026.

\paragraph{Novel-State Synthesis}
Table~\ref{tab:partnet_joints} reports joint estimation results on PartNet-Mobility. DailyArt achieves the best Overall Success Rate of 68.4, surpassing the strongest baseline, Physx-Anything (62.8), by 5.6 points. The improvement is also reflected in all individual joint attributes: the Type error decreases to 0.215, the Origin error to 0.124, the Direction error to 0.275, and the Range error to 0.242. These results suggest that the proposed synthesis-mediated formulation improves joint estimation as a whole, rather than benefiting only a single attribute.

A similar trend is observed on AKB-48 in Table~\ref{tab:akb_joints}. DailyArt again achieves the best Overall Success Rate at 54.4, compared with 52.8 for Physx-Anything and 48.3 for Articulate-Anything. It also yields the lowest Type, Direction, and Range errors, while matching the best Origin error at 0.204. Since AKB-48 consists of real-world objects evaluated in a cross-domain setting, these results indicate that the proposed formulation transfers beyond the synthetic benchmark while maintaining strong overall joint estimation performance.

\begin{table}[t]
\centering
\caption{\textbf{PartNet-Mobility joint estimation.} We report the Overall Success Rate (\%, $\uparrow$) and mean errors ($\downarrow$) for individual joint attributes.}
\vspace{-3mm}
\label{tab:partnet_joints}
\setlength{\tabcolsep}{4.0pt}
\resizebox{\linewidth}{!}{
\begin{tabular}{l | c | c c c c} 
\toprule
\textbf{Method} & \textbf{Overall} $\uparrow$ & \textbf{Type} $\downarrow$ & \textbf{Origin} $\downarrow$ & \textbf{Direct.} $\downarrow$ & \textbf{Range} $\downarrow$ \\
\midrule
URDFormer \cite{chen2024urdformer} & 48.6 & 0.342 & 0.188 & 0.370 & 0.335 \\
Singapo \cite{liu2024singapo} & 35.4 & 0.482 & 0.285 & 0.512 & 0.440 \\
Articulate-Anything \cite{le2025articulateanything} & 56.0 & 0.288 & 0.165 & 0.322 & 0.310 \\
Physx-Anything \cite{cao2025physx} & 62.8 & 0.295 & 0.130 & 0.325 & 0.282 \\
\midrule
\rowcolor{grey!10} \textbf{DailyArt (Ours)} & \textbf{68.4} & \textbf{0.215} & \textbf{0.124} & \textbf{0.275} & \textbf{0.242} \\
\bottomrule
\end{tabular}
}
\vspace{-3mm}
\end{table}

\begin{table}[t]
\centering
\caption{\textbf{AKB-48 joint estimation.} We report the zero-shot Overall Success Rate (\%, $\uparrow$) and mean errors ($\downarrow$) for individual joint attributes.}
\vspace{-3mm}
\label{tab:akb_joints}
\setlength{\tabcolsep}{3.8pt}
\resizebox{\linewidth}{!}{
\begin{tabular}{l | c | c c c c} 
\toprule
\textbf{Method} & \textbf{Overall} $\uparrow$ & \textbf{Type} $\downarrow$ & \textbf{Origin} $\downarrow$ & \textbf{Direct.} $\downarrow$ & \textbf{Range} $\downarrow$ \\
\midrule
URDFormer \cite{chen2024urdformer} & 37.5 & 0.738 & 0.395 & 0.625 & 0.482 \\
Singapo \cite{liu2024singapo} & 32.4 & 0.819 & 0.372 & 0.584 & 0.466 \\
Articulate-Anything \cite{le2025articulateanything} & 48.3 & 0.370 & 0.268 & 0.351 & 0.403 \\
Physx-Anything \cite{cao2025physx} & 52.8 & 0.338 & 0.204 & 0.377 & 0.371 \\
\midrule
\rowcolor{grey!10} \textbf{DailyArt (Ours)} & \textbf{54.4} & \textbf{0.275} & \textbf{0.204} & \textbf{0.349} & \textbf{0.368} \\
\bottomrule
\end{tabular}
}
\vspace{-3mm}
\end{table}

\begin{table}[t]
\centering
\caption{\textbf{PartNet-Mobility novel-state synthesis.} We report the visual fidelity and semantic consistency of the synthesized opened state $\hat{\mathbf{I}}_{1}$.}
\vspace{-3mm}
\label{tab:partnrt_synthesis}
\setlength{\tabcolsep}{3.5pt}
\resizebox{0.95\linewidth}{!}{
\begin{tabular}{l|ccccc}
\toprule
\textbf{Method} & \textbf{PSNR} $\uparrow$ & \textbf{SSIM} $\uparrow$ & \textbf{LPIPS} $\downarrow$ & \textbf{CLIP-T} $\uparrow$ & \textbf{FVD} $\downarrow$ \\
\midrule
DragAPart \cite{li2024dragapart} 
& 21.2 & 0.837 & 0.143 & 0.632 & 212.4 \\
PartRM \cite{gao2025partrm} 
& 22.8 & 0.840 & 0.145 & 0.643 & 219.5 \\
Puppet-Master \cite{li2025puppet} 
& 23.8 & 0.829 & 0.110 & 0.678 & 204.3 \\
LARM \cite{yuan2025larm} 
& 24.3 & 0.907 & 0.104 & 0.749 & 205.4 \\
\midrule
\rowcolor{grey!10} \textbf{DailyArt (Ours)} 
& \textbf{25.5} & \textbf{0.920} & \textbf{0.102} & 0.766 & \textbf{202.2} \\
\bottomrule
\end{tabular}
}
\vspace{-3mm}
\end{table}
\begin{table}[t]
\centering
\caption{\textbf{AKB-48 novel-state synthesis.} We report the zero-shot quality of the synthesized opened state $\hat{\mathbf{I}}_{1}$ on real-world objects.}
\vspace{-3mm}
\label{tab:akb_synthesis}
\setlength{\tabcolsep}{3.5pt}
\resizebox{0.95\linewidth}{!}{
\begin{tabular}{l|ccccc}
\toprule
\textbf{Method} & \textbf{PSNR} $\uparrow$ & \textbf{SSIM} $\uparrow$ & \textbf{LPIPS} $\downarrow$ & \textbf{CLIP-T} $\uparrow$ & \textbf{FVD} $\downarrow$ \\
\midrule
DragAPart \cite{li2024dragapart} & 16.3 & 0.724 & 0.355 & 0.512 & 312.4 \\
PartRM \cite{gao2025partrm} & 18.1 & 0.752 & 0.285 & 0.523 & 268.5 \\
Puppet-Master \cite{li2025puppet} & 17.8 & 0.815 & 0.249 & 0.534 & 246.1 \\
LARM \cite{yuan2025larm} & 18.3 & 0.813 & 0.174 & 0.654 & 265.4 \\
\midrule
\rowcolor{grey!10} \textbf{DailyArt (Ours)} & \textbf{19.6} & \textbf{0.821} & \textbf{0.162} & 0.656 & \textbf{245.2} \\
\bottomrule
\end{tabular}
}
\vspace{-3mm}
\end{table}

\paragraph{Joint Estimation}
Table~\ref{tab:partnrt_synthesis} summarizes novel-state synthesis results on PartNet-Mobility. DailyArt obtains the strongest overall performance across all reported metrics, reaching 25.5 PSNR, 0.920 SSIM, 0.102 LPIPS, 0.766 CLIP-T, and 202.2 FVD. Compared with the strongest competing baseline, this corresponds to gains of +1.2 PSNR, +0.013 SSIM, -0.002 LPIPS, +0.017 CLIP-T, and -2.1 FVD. These results show that the synthesized opened states are both visually faithful and semantically consistent with the intended articulation, supporting the role of Stage~I as an effective intermediate for downstream joint reasoning.

The same pattern holds on the zero-shot AKB-48 benchmark in Table~\ref{tab:akb_synthesis}. DailyArt improves PSNR from 18.3 to 19.6, SSIM from 0.815 to 0.821, LPIPS from 0.174 to 0.162, CLIP-T from 0.654 to 0.656, and FVD from 246.1 to 245.2. Although the gains are smaller than those on joint estimation, they are consistent across metrics and datasets, suggesting that the synthesis module remains reliable under more challenging real-world conditions.


\subsection{Ablation Studies}
\label{sec:ablation}

Table~\ref{tab:ablation_main} studies the main design choices in DailyArt. We first examine the necessity of the two-stage pipeline, and then analyze several module-level design choices.

\begin{figure*}[t]
    \centering
    \includegraphics[width=0.95\linewidth]{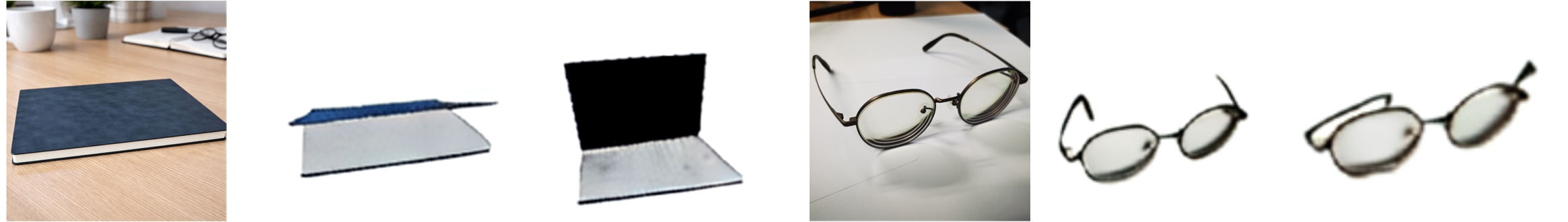} 
    \caption{\textbf{Qualitative ablation results.} Left: Without Stage 1 target-state synthesis, direct joint regression from a single closed-state image often fails to synthesise a plausible articulation state (the opened notebook is more like a laptop). Right: Without 3D lifting, a 2D pair encoder may appear reasonable in the image plane but produces incorrect joint geometry in 3D (inaccurate joint estimations could affect the kinematic part's motion). These examples highlight the importance of both synthesis-mediated cross-state reasoning and 3D geometric constraints in DailyArt.}
    \label{fig:exp_ablation_vis}
\end{figure*}

\paragraph{Necessity and Reliability of Target State Synthesis}
Rows \textbf{A} and \textbf{B} evaluate the role of target-state synthesis in the overall pipeline. In Row \textbf{A}, we remove Stage 1 and directly regress joint parameters from the input image. This reduces the Overall Success Rate from 68.4\% to 44.2\%, indicating that direct single-image regression is substantially more difficult than synthesis-mediated estimation. Figure~\ref{fig:exp_ablation_vis} (left) visually confirms this: without target-state synthesis, the predicted articulation often severely misaligns with the object's actual movable structure. In Row \textbf{B}, we replace the synthesized target state with the ground-truth opened state rendered by the simulator. This oracle setting reaches 69.7\% Overall Success Rate, which is only 1.3\% above the full model. This minimal gap demonstrates that our synthesized target states are highly reliable and provide sufficient geometric cues for accurate downstream joint estimation.

\begin{table}[t]
\centering
\caption{\textbf{Ablation Study.} We first validate the core two-stage pipeline design (I), and then study several design choices in the full model (II). The \textit{Full} model provides the best overall balance among the non-oracle configurations evaluated here.}
\vspace{-3mm}
\label{tab:ablation_main}
\setlength{\tabcolsep}{4pt}
\resizebox{\linewidth}{!}{
\begin{tabular}{c|l|ccc}
\toprule
& \textbf{Configuration} & \textbf{PSNR} $\uparrow$ & \textbf{Overall SR} $\uparrow$ & \textbf{Latency} $\downarrow$ \\
\midrule
\rowcolor{grey!10} & \textbf{DailyArt (Full Pipeline)} & \textbf{25.5} & \textbf{68.4\%} & \textbf{0.45s} \\
\midrule
\multicolumn{5}{l}{\textbf{I. Pipeline}} \\
\midrule
\multirow{2}{*}{\textbf{A}} & Direct Regression (No Synthesis) & -- & 44.2\% & \textbf{0.18s} \\
 & \textit{\small Change vs. Full} & \textit{--} & \textit{-24.2\%} & \textit{-0.27s} \\
\midrule
\multirow{2}{*}{\textbf{B}} & Oracle Synthesis (GT Target State) & -- & \textbf{69.7\%} & 0.20s \\
 & \textit{\small Gap to Upper Bound} & \textit{--} & \textit{+1.3\%} & \textit{--} \\
\midrule
\multicolumn{5}{l}{\textbf{II. Module Design}} \\
\midrule
\multirow{2}{*}{\textbf{C}} & w/ 2D Pair-Encoder (No 3D Lifting) & 25.5 & 50.0\% & 0.38s \\
 & \textit{\small Change vs. Full} & \textit{--} & \textit{-18.4\%} & \textit{-0.07s} \\
\midrule
\multirow{2}{*}{\textbf{D}} & w/ Sequential Generation & 25.0 & 64.8\% & 1.45s \\
 & \textit{\small Change vs. Full} & \textit{-0.5} & \textit{-3.6\%} & \textit{+1.00s} \\
\bottomrule
\end{tabular}
}
\vspace{-3mm}
\end{table}

\paragraph{Role of 3D Lifting}
Row \textbf{C} isolates the contribution of the 3D lifting module. Replacing it with a 2D pair encoder leaves image synthesis quality unchanged, but reduces the Overall Success Rate from 68.4\% to 50.0\%. This result suggests that high-quality synthesized images alone are not enough for precise joint estimation, and that 3D geometric reasoning plays an important role in converting cross-state differences into reliable kinematic predictions. As shown in Figure~\ref{fig:exp_ablation_vis} (right), the predictions from the 2D pair encoder may appear plausible in the image plane, but catastrophic errors become evident under side views, where the estimated joint geometry is no longer consistent in 3D.

\paragraph{Synthesis Strategy}
Row \textbf{D} compares direct target-state synthesis with sequential generation. Sequential generation reduces PSNR from 25.5 to 25.0 and Overall Success Rate from 68.4\% to 64.8\%, while increasing latency from 0.45s to 1.45s. This confirms that our direct synthesis strategy is both more accurate and significantly more efficient for this task.

\section{Conclusion}
\label{sec:conclusion}
We presented \textbf{DailyArt}, a synthesis-first pipeline that enables \emph{single-image} articulation understanding by converting static closed-state perception into cross-state discrepancy reasoning between an observed input and a synthesized open-state counterpart. 
DailyArt is built on two technical contributions: \textbf{(i) novel state synthesis}, where the articulation index $t$ is injected via \emph{AdaLN-based global modulation} to stably produce large-deformation target states without test-time masks or oracle priors; and \textbf{(ii) joint estimation}, where we lift the image pair into 3D point-maps and identify motion-seed cues from spatial displacement to ground joint inference in an object-centric geometry under occlusion and depth-dependent axes. 
Across synthetic benchmarks and diverse real-world images, DailyArt improves joint parameter accuracy and category-level generalization over prior single-image methods, while narrowing the gap to approaches that rely on real state transitions. More broadly, since DailyArt operates purely from image observations, it may potentially benefit world models and embodied environments that require joint cues in offline simulations before on-device interaction.  

\begin{figure}[t]
    \centering
    \includegraphics[width=0.95\linewidth]{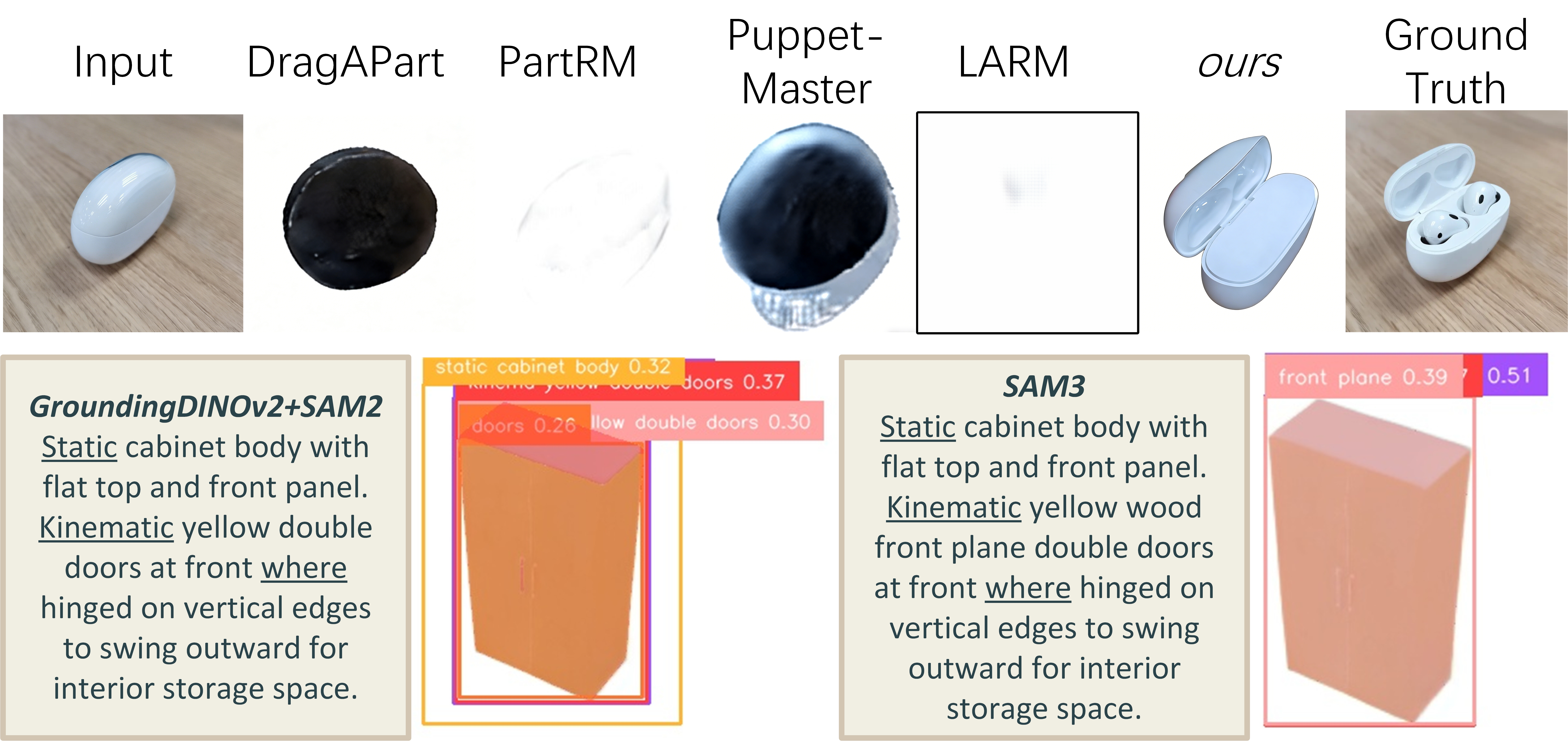} 
    \caption{\textbf{Failure cases on real-world unseen object and part segmentation on articulated objects.} Even when conditioned with optimal text or point prompts, off-the-shelf foundation segmentation models fail to separate moving parts from the static base. This suggests that priors and prompts are less effective for identifying kinematic structures than we thought.}
    \label{fig:conclu_failure}
    \vspace{-3mm}
\end{figure}

\paragraph{Limitations and future work} Current limitations stem from the reliance on novel state synthesis fidelity and discretized state modeling, where synthesis errors may propagate to later joint estimation. Current baselines or frameworks may not support extreme articulations in industrial areas. And 3D lifting used in DailyArt may fail when none of the motions is available (objects facing backwards), which is also difficult for human beings to identify any articulation. 
In addition, DailyArt assumes that the object admits a well-defined closed state and a canonical maximally-open target configuration. Yet there are articulated objects without clear endpoint states that may violate this assumption and degrade cross-state correspondence. Notably, as illustrated in Fig.~\ref{fig:conclu_failure}, the difficulty of foundation segmentation models to delineate parts using static semantics alone reinforces our core premise: a synthesized novel state provides an better assistant for joint estimation instead of word descriptions using LLMs or image annotations from human.

{
    \small
    \bibliographystyle{IEEEtran}
    \bibliography{sample-bibliography}
}
\newpage
\begin{figure*}[h]
    \centering
    \includegraphics[width=0.85\textwidth]{figures/image_page1.jpg}
    \caption{{\bf Visual Comparison on Joint-conditioned Novel State Synthesis (Stage III)} of DailyArt and baselines. We prepared priors for baselines, such as drags (calculated from input and gt meshs), seg masks from LLM, and camera extrinsic.}
    \label{fig:page1}
\end{figure*}
\newpage
\begin{figure*}[h]
    \centering
    \includegraphics[width=0.85\textwidth]{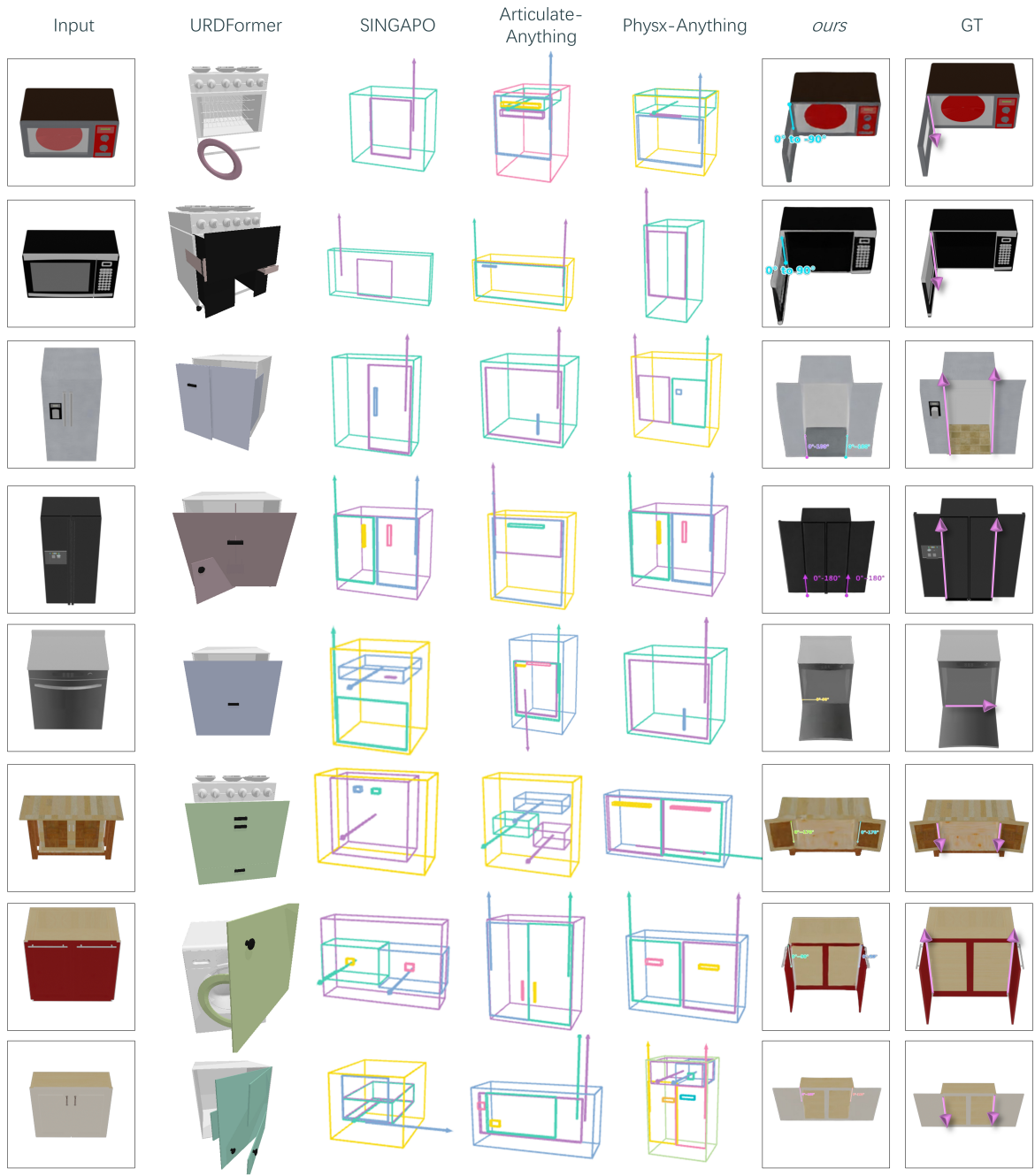}
    \caption{{\bf Visual Comparison on Joint Estimation (Stage II).} The visualization results differ due to the variations on how methods predict the joint parameters, including engine annotations, mesh building files or 3D coordinates. Instead of generating only a URDF structure for the simulation engine or part retrievals, DailyArt estimates joints based on the current view, and provides part control information possible for both generative pipelines or robot interactions, including motion ranges and axis directions.}
    \label{fig:page1}
\end{figure*}

\end{document}